\documentclass[letterpaper, 10 pt, journal, twoside]{ieeetran}

\usepackage{float}
\usepackage{graphics} % for pdf, bitmapped graphics files
\usepackage{epsfig} % for postscript graphics files
\usepackage{mathptmx} % assumes new font selection scheme installed
\usepackage{times} % assumes new font selection scheme installed
\usepackage{amssymb}  % assumes amsmath package installed
\usepackage{amsmath,amsfonts}
\usepackage{mathtools}
\usepackage{physics}
\usepackage{algorithmicx}
\usepackage{algorithm}
\usepackage{makecell}
\usepackage{algpseudocode}
\usepackage{array}
\usepackage{textcomp}
\usepackage{stfloats}
\usepackage{url}
\usepackage{verbatim}
\usepackage{graphicx}
\usepackage{cite}
\usepackage{adjustbox}
\usepackage{tabularx}
\usepackage{booktabs}
\usepackage{multirow}
\usepackage{wasysym}
\usepackage{diagbox}
\usepackage[table,xcdraw]{xcolor} 
\usepackage{siunitx}

\definecolor{ce}{RGB}{204,255,204}
\definecolor{gt}{RGB}{204,229,255}

\usepackage[acronym,nolist]{glossaries} % “nolist” = do not print a list
\makeglossaries 
\newacronym{dof}{DoF}{degrees of freedom}
\newacronym{ft}{F/T}{force/torque}
\newacronym{gf}{GF}{grasp force}
\newacronym{ifea}{iFEA}{inverse finite element analysis}
\newacronym{fea}{FEA}{finite element analysis}
\newacronym{ros}{ROS}{robot operating system}
\newacronym{rmse}{RMSE}{root mean square error}
\newacronym{qp}{QP}{Quadratic Programming}
\newacronym{tpu}{TPU}{thermoplastic polyurethane}
\newacronym{sofa}{SOFA}{Simulation Open Framework Architecture}
\newacronym{gl}{GL}{left gripper}
\newacronym{gr}{GR}{right gripper}
\newacronym{dlc}{DLC}{Deeplabcut}
\newacronym{pcd}{PCD}{point-cloud data}
\newacronym{nrmsd}{NRMSD}{normalized root mean square deviation}
\newacronym{std}{STD}{standard deviation}
\newacronym{me}{ME}{maximum error}
\newacronym{icp}{ICP}{iterative closest point}

\def\BibTeX{{\rm B\kern-.05em{\sc i\kern-.025em b}\kern-.08em
    T\kern-.1667em\lower.7ex\hbox{E}\kern-.125emX}}
    
\usepackage{etoolbox} 
\usepackage{tikz}
\newcommand\copyrightnotice[1]{
    \begin{tikzpicture}[remember picture,overlay]
    \node[anchor=south,yshift=12pt] at (current page.south) {\fbox{\parbox{\dimexpr\textwidth-\fboxsep-\fboxrule\relax}{#1}}};
    \end{tikzpicture}
}

\begin{document}

\bstctlcite{IEEEexample:BSTcontrol}

\title{\LARGE \bf A Model-based Visual Contact Localization and Force Sensing System for Compliant Robotic Grippers}

%only for final version
\author{Kaiwen Zuo$^{1}$, Shuyuan Yang$^{1}$ and Zonghe Chua$^{1}$
\thanks{Manuscript received: February 5, 2026; Revised:
March 19, 2026; Accepted: April 25, 2026.}
\thanks{This paper was recommended for publication by
Editor Ashis Banerjee upon evaluation of the Associate Editor and Reviewers’
comments.}
\thanks{This work is supported by the Office of Naval Research (ONR) Award N00014-23-1-2842.}
\thanks{$^{1}$K.\,Zuo, S.\,Yang and Z.\,Chua are with the Department of Electrical, Computer, and Systems Engineering, Case Western Reserve University, Cleveland, OH 44106, USA (e-mail: kxz365@case.edu; sxy841@case.edu; zxc703@case.edu)
}%
\thanks{Digital Object Identifier (DOI): see top of this page.}
}

\markboth{IEEE Robotics and Automation Letters. Preprint Version. Accepted April, 2026}{Zuo \MakeLowercase{\textit{et al.}}: A Model-based Visual Contact Localization and Force Sensing System for Compliant Robotic Grippers} 

\maketitle

%%%%%%%%%%%%%%%%%%%%%%%%%%%%%%%%%%%%%%%%%%%%%%%%%%%%%%%%%%%%%%%%%%%%%%%%%%%%%%%%
\begin{abstract}

Grasp force estimation can help prevent robots from damaging delicate objects during manipulation and improve learning-based robotic control. Integrating force sensing into deformable grippers negotiates trade-offs in cost, complexity, mechanical robustness, and performance. With the growing integration of RGB-D wrist cameras into robotic systems for control purposes, camera-based techniques are a promising solution for indirect visual force estimation. Current approaches mostly utilize end-to-end deep learning, which can be brittle when generalizing to new scenarios, while existing model-based approaches are unsuited to grasping and modern grasper geometries. To address these challenges, we developed a model-based visual force sensing approach integrating an iterative contact localization with generalization to unseen objects. The system extracts structural key points from wrist camera RGB-D images of deforming fin-ray-shaped soft grippers, and uses these key points to define parameters of an \acrlong{ifea} simulation in \acrlong{sofa}. The iterative contact localization sub-system utilizes a deep learning-based online 3D reconstruction and pose estimation pipeline to dynamically update contact location, and is robust to visual occlusion and unseen objects. Our system demonstrated an average \acrlong{rmse} of 0.23\,N and \acrlong{nrmsd} of 2.11\% during the load phase, and 0.48\,N and 4.34\% over the entire grasping process when interacting with different objects under various conditions, showcasing its potential for real-time model-based indirect force sensing of soft grippers.

\end{abstract}

%%%%%%%%%%%%%%%%%%%%%%%%%%%%%%%%%%%%%%%%%%%%%%%%%%%%%%%%%%%%%%%%%%%%%%%%%%%%%%%%
\section{Introduction}

Soft grippers are widely used in robotic manipulation due to their inherent safety and adaptability. These properties, together with their high mechanical robustness, make them widely used in learning-based dexterous manipulation. In this application, providing training demonstrators and the control policy access to grasp force information has been shown to enhance autonomous manipulation performance, enabling precise and reliable interactions with target objects \cite{cuan2024leveraging, abdelaal2025force}. Integrating force sensing into compliant grippers requires negotiating trade-offs in cost, sensor size, sensing range, accuracy, reliability, and integration complexity. Metal strain-gauge sensors are susceptible to damage when exposed to impacts. Integrating a force sensor into a soft gripper without disrupting its natural deformation behavior during grasping \cite{integrate_2} is also challenging. Novel force sensors, such as distributed soft tactile sensors\cite{tactile_review, revier5_cite} and discrete bend sensors\cite{tro_fr_dflct}, preserve the gripper's natural deformation. However, the distributed tactile sensors often suffer from the difficulty of decoupling external contact forces from the gripper’s intrinsic deformation, while discrete sensors typically have constrained sensing regions due to the physical limitation of their mechatronic design.

To address this challenge, indirect sensing approaches, such as camera-based force sensing, have been developed \cite{jc_2,tro_fr_rob,vison_dl,vision_science, zebrafish, reviewer3_cite}.  These approaches use RGB-D images to observe gripper deformation arising from external interaction, and are advantageous because of their relatively low cost, easy accessibility, and unobtrusiveness. Such an approach is further suited for modern learning-based dexterous manipulation, as most setups already require wrist cameras \cite{intelligence2025pi_}. 

Data-driven approaches have been widely adopted for indirect visual force estimation because of their ability to learn complex representational mappings. These have typically used a neural network to observe the gripper deformation \cite{jc_2, tro_fr_rob}, with some using finite-element simulation to create synthetic training data \cite{vison_dl}. These studies reveal a shortcoming of data-driven methods, which is their dependency on the quality and diversity of training data, which can significantly limit their generalizability. Moreover, the performance of such approaches relies on the quality of the visual features observed from the gripper. To ensure accuracy, past works have positioned the camera parallel to the gripper's main deformation plane, which can significantly limit the robot’s workspace in real-world applications \cite{vision_science, tro_fr_rob, vison_dl}. Alternatively, mounting the camera on the robot’s wrist increases the feasible workspace but typically results in reduced estimation accuracy \cite{jc_2}.

Model-based approaches, such as \acrfull{ifea}, can provide more robust out-of-distribution performance. Reddy et al. \cite{zebrafish} estimated both the actuation and the grasp force on miniature compliant grippers by solving Cauchy’s problem in elasticity. However, their method cannot be implemented in real time due to its stated computational inefficiency. Zhang et al. \cite{sofarobot} addressed this limitation by using reduced-order approximations and a \acrfull{qp}-based solver in \acrfull{sofa} \cite{realtimeqp}. Their approach focused on estimating external forces acting on a soft robot with \textit{visible} contact locations. However, when grasps are viewed from an angled wrist camera, contact locations are often occluded.

%%%%%%%%%%%%%%%%%%%%%%%%%%%%%%%%%%%%%%%%%%%%%%%%%%%%%%%%%%%%%%%%%%
% Copyright for arxiv
\copyrightnotice{\scriptsize \copyright \, 2026 IEEE.  Personal use of this material is permitted.  Permission from IEEE must be obtained for all other uses, in any current or future media, including reprinting/republishing this material for advertising or promotional purposes, creating new collective works, for resale or redistribution to servers or lists, or reuse of any copyrighted component of this work in other works.}

In this letter, we present an approach to address the limitations of existing camera-based methods, including their high sensitivity to extracted features and the requirement for visibility of contact locations. We apply real-time \acrshort{ifea} simulations using \acrshort{sofa} to the problem of grasp force acquisition for fin-ray-shaped soft grippers, specifically dealing with non-visible contact interactions. Our approach introduces the following contributions:

{\setlength{\topsep}{1em}%
\setlength{\partopsep}{0pt}%
\setlength{\parsep}{0pt}%
\setlength{\itemsep}{1em}
\begin{enumerate}
    \item A novel image-based mesh reconstruction approach to resolve the scale mismatch between the physical object and a learning-based model-generated mesh.
    \item A novel model-based iterative contact localization algorithm capable of handling occluded contact locations and pre-estimating the contact location.
    \item Experimentally validated accuracy through both static and on-robot experiments, supported by further demonstrations of the usefulness and effectiveness of the contact estimator.
\end{enumerate}}

\begin{figure*}[!t]
    \vspace{.25em}
    \centering
    \includegraphics[width=\linewidth]{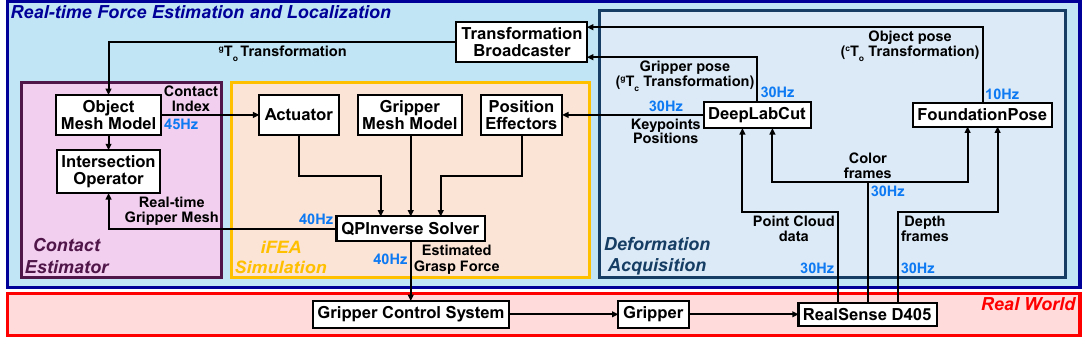}
    \caption{Flow chart for the contact localization and force sensing system pipeline with the corresponding frame rates for the key components.}
    \label{Fig:flow_chart}
\end{figure*}

\section{Materials and Method}
\subsection{System Overview}

Two fin-ray-shaped soft gripper jaws were mounted to an electromechanical base and observed by a wrist-mounted RGB-D camera. The structural key points of the grippers are segmented from the RGB-D image stream using \acrfull{dlc}~\cite{DLC}, a deep neural-network-based skeletal pose estimation framework that leverages transfer learning. Their 3D positions are input to \acrshort{ifea} simulations developed using \acrshort{sofa}. The image stream is also used in conjunction with a reconstructed mesh of the grasped object to localize its pose relative to the gripper jaws via FoundationPose \cite{foundationposewen2024}. Using the object and gripper mesh intersection, a model-based iterative contact localization algorithm computes the contact position between the grasped object and the gripper. This contact estimate is provided to the simulation, and the grasp force is then computed based on the acquired constraints. The flow chart of the system pipeline and the corresponding frame rate of the key components are shown in Fig.\,\ref{Fig:flow_chart}.

\subsection{Gripper Geometry and Fabrication}
\begin{figure}[!b]
\centering
\includegraphics[width=\linewidth]{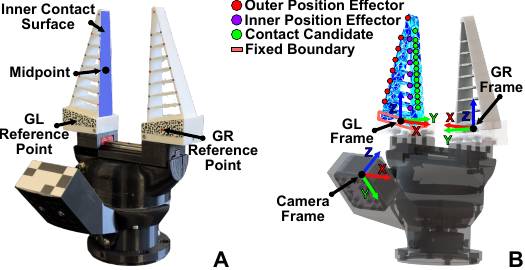}
\caption{Physical dual-jaw gripper and its digital twin. (A) Physical dual-jaw gripper with the RGB-D camera mounted on its wrist. (B) CAD model of the dual-jaw gripper, the left gripper is overlaid with the \acrshort{ifea} simulation mesh, and its relevant simulation boundary conditions. The outer and inner position effectors and contact candidates are indicated. }
\label{Fig:kps_demo}
\end{figure}

Two fin-ray-shaped grippers were 3D-printed using \acrfull{tpu} 95A. \acrshort{tpu} 95A was used because it is commercially available, and has a relatively large linear region in its stress-strain curve \cite{tpu_characterization}. The latter property made it compatible with the \acrshort{qp} Inverse Solver from the SofaRobots Inverse plugin \cite{realtimeqp}, which was developed under the assumption of linear elasticity. The geometry and size of our gripper are similar to those of the Festo adaptive gripper (DHAS-GF-80-U-BU), but with thinner bands and ribs to achieve greater compliance. The effective compliance was evaluated by displacing a 20mm-diameter cylinder 10\,mm at the gripper midpoint,  and showed our gripper achieving two times greater compliance than the Festo gripper. We installed the grippers on an SSG-48 adaptive electric gripper base (Source Robotics, Zagreb, Croatia) as shown in Fig.\,\ref{Fig:kps_demo}A. 

\subsection{SOFA Simulation Setup} \label{section:sofa setup}

Left gripper (\acrshort{gl}) refers to the gripper jaw located on the left side of the camera view, and \acrfull{gr} refers to the one on the right (Fig.\,\ref{Fig:kps_demo}B). Both grippers use a mesh with 522 vertices to balance simulation accuracy and computational cost. The simulations for each gripper run in parallel, enabling a simulation frame rate up to 40 Hz. The simulation takes key point positions and contact location as inputs and outputs the estimated grasp force (Fig.\,\ref{Fig:flow_chart}).

To build the simulation in \acrshort{sofa}, we define two constraints: position effector constraints and an actuator constraint. In each iteration of the simulation, the \acrshort{qp} Inverse Solver will compute the force applied to the gripper at the actuator position that minimizes the distance between the position effectors and their corresponding targets. As shown in Fig.\,\ref{Fig:kps_demo}B, we chose the nodes that correspond to structural key points on the physical gripper to be the position effectors, and they are categorized as outer or inner effectors based on their positions with respect to the inner contact surface. These key points are thus the targets of their respective position effectors and are captured by our key point acquisition pipeline as described in Sec.\,\ref{section:dlc}.

For the actuator constraint, we chose to use the force point actuator rather than the force surface actuator because it provided more stable solutions during preliminary tests. The position of the force point actuator is crucial to the simulation, as an incorrect contact condition can lead to inaccurate force estimation. \acrshort{sofa} allows us to initialize the actuator constraint with a set of contact candidates. The actuator constraint can be dynamically changed by mounting the actuator on a different contact candidate during simulation. Here, we selected 14 linearly arranged points along the central axis of the inner contact surface as contact candidates whose positions in the undeformed state are denoted as $\{ \mathbf{a}_i \}$, and the actuator is initialized on the middle candidate.

Another key factor in the simulation is $\varepsilon$, which represents the weight of the deformation energy term in the QP inverse solver objective function \cite{realtimeqp}, given as
\begin{equation}
\label{Eq: qpepsilon}
%\min\Bigl\{ 
J=
\tfrac12\,\lambda_a^{\top}\!\bigl(W_{ea}^{\top}W_{ea}+\varepsilon W_{aa}\bigr)\lambda_a
+\lambda_a^{\top}W_{ea}\,\delta_e^{\mathrm{free}}
%\Bigr\}
\text{.}
\end{equation}
The solver accounts for more internal strain and less shape transformation with higher $\varepsilon$. According to the datasheet of the Festo gripper, the effective stiffness of the fin-ray-shaped gripper varies at different contact positions, indicating that the ratio of internal strain and shape transformation is dependent on where contact occurs. Therefore, $\varepsilon$ was empirically tuned for each contact candidate. At each iteration, the actuator is initialized on the estimated contact candidate, and the corresponding $\varepsilon$ value is updated.

\subsection{Key Points Acquisition and Calibration}\label{section:dlc}

3D positions of structural key points were identified using \acrshort{dlc} \cite{DLC} from RGB-D data captured via an Intel RealSense D405 at 30\,Hz. 
\acrshort{dlc}  requires only minimal labeled data ($\sim$50-200 frames) to adapt to a specific skeletal pose estimation task. 
% To adopt \acrshort{dlc} to our application, 
We collected and annotated 300 frames of the gripper grasping the cylinder and the cube used in our on-robot evaluation. We labeled 15 structural key points and one reference point per gripper in each frame. When the key point is occluded, the corresponding label remains unannotated.
Reference points as defined in Fig.\,\ref{Fig:kps_demo}A were used to localize the gripper jaw during the grasping process. These data were used to fine-tune a pretrained \acrshort{dlc} model using the official API, which allows adjusting the output dimensionality without directly modifying the network architecture.

In each frame of the camera stream, \acrshort{dlc} predicts the pixel indices of the key points and reference points in RGB images with their corresponding likelihood. Then, we obtain the position data of the key points $\{ \prescript{\text{c}}{}{\mathbf{q}}_{\text{kp}, i}\}$ and reference point $\prescript{\text{c}}{}{\mathbf{q}}_{\text{rp}}$ in the camera frame for each gripper by relating the pixel indices to the point cloud data. The key points are mapped from the camera to the gripper jaw frame using the transformation matrix $\prescript{\text{g}}{}{\mathbf{T}}_{\text{c}}$, which is computed by chaining the camera-to-gripper-base transformation $\prescript{\text{r}}{}{\mathbf{T}}_{\text{c}}$, and the gripper-base-to-jaw transformation $\prescript{\text{g}}{}{\mathbf{T}}_{\text{r}}$. The gripper base frame shares the orientational alignment as the jaw frame, but its origin tracks the reference point, which translates as the gripper jaws open and close. $\prescript{\text{g}}{}{\mathbf{T}}_{\text{r}}$ and $\prescript{\text{r}}{}{R}_{\text{c}}$, the rotational component of $\prescript{\text{r}}{}{\mathbf{T}}_{\text{c}}$, are specified from CAD models.

Capturing all the key points allows for more constraints to be specified in the simulation, which can improve the system accuracy. However, some key points are occluded or may have incorrect position estimates due to misidentification in \acrshort{dlc} or noisy point cloud reconstruction in most grasping scenarios. These can lead to infeasible position effector target specifications that can contribute to inaccurate and unstable solutions. To prevent this, we applied key point detection confidence thresholding followed by spatial bounding box filtering for all key points. The threshold confidence is determined empirically based on the model output to coarsely retain most visible key points while filtering out all the occluded ones. We set the spatial bounding box to contain the workspace for key points to filter out the incorrectly inferred ones in the background.

When initializing the position effector targets at each simulation iteration, the simulation controller receives an indexed list of incorrect key points and deactivates their corresponding position effector. We use a binary switch instead of continuous smoothing, since occluded key points can drift unpredictably onto the object surface or into the background, which makes the continuous smoothing unreliable.

\subsection{Contact Estimator}\label{section:cnt estimation}

The contact position serves as an essential constraint in the simulation. Unlike in existing approaches \cite{sofarobot}, our perspective camera view obscures the contact position in the point cloud data. Additionally, the simulated gripper deformation and force depend on accurate contact localization, which in turn depends on the gripper deformation. This phenomenon will be more pronounced with larger deformation and will reduce the performance of naive closest point approaches \cite{sofarobot}. To address these, we developed a novel model-based iterative contact localization algorithm. Since pose estimation is not the primary contribution of this work, we use FoundationPose \cite{foundationposewen2024} to estimate the object pose, by providing it with a reconstructed mesh of the grasped object. This object mesh is also used as a digital twin, whose pose is described in the frame of the deformed gripper jaw.

\subsubsection{Mesh Reconstruction}
The object mesh is initially generated using the SAM-3D Objects model\cite{sam3d}. Given an RGB image and the object mask, the model reconstructs the object mesh in a canonical space and jointly estimates the object scale and pose. Because the model does not take the camera intrinsics as input, there is a severe scale mismatch between the mesh and the physical object, leading to inaccurate pose estimation and contact localization. To address these, we developed a novel scale calibration workflow.

Using the RGB-D camera, we segment the object point cloud from the point cloud data with the object mask. The reconstructed mesh is first scaled and transformed into the camera frame based on the model's output. The mesh is then coarsely scaled using the x-axis bounding box of the acquired point cloud. However, relying solely on coarse calibration lacks robustness, as it depends on accurate pose estimation. 

Since the mesh represents the complete object geometry, whereas the acquired point cloud only contains partial observations, directly applying the iterative closest point(\acrshort{icp}) algorithm yields poor alignment. To obtain the partial view of the mesh, we project it onto a pixel grid defined on the XY plane of the camera frame, with the origin aligned to the camera frame origin, and the grid resolution determined by the camera intrinsics. We obtain a partial mesh by preserving the points closest to the origin for each pixel. Then we perform \acrshort{icp} twice between the partial mesh and the acquired point cloud using Open3D, enabling scale adjustment on the second \acrshort{icp}. We apply the resulting scale factor to the complete reconstructed mesh and generate a watertight triangular mesh via Poisson surface reconstruction.
\begin{figure}[!t]
\vspace{-0.75em}
\begin{algorithm}[H]
\small
% \footnotesize
\caption{Iterative Contact Localization Algorithm}
\label{Alg:cntest}
\begin{algorithmic}[1] 
\State \textbf{Input:} undeformed vertex matrix $\mathbf{V}$, object mesh $\mathbf{M}_\text{o}$, candidate set $\{\mathbf{a}_i\}$, offset $l_0$, contact status flag \textbf{S} = False
\For{each \acrshort{ifea} simulation iteration}
    \State Compute $(\prescript{\text{g}}{}{R}_{\text{o}}, \prescript{\text{g}}{}{t}_{\text{o}}) \gets (\prescript{\text{g}}{}{\mathbf{T}}_{\text{c}} , \prescript{\text{c}}{}{\mathbf{T}}_{\text{o}})$
    \State Get $\overline{\mathbf{V}}$ and reconstruct $\overline{\mathbf{M}}_g \gets \overline{\mathbf{V}}$
    \State $\hat{\mathbf{M}}_\text{o} \gets \mathbf{M}_\text{o}$, and apply rotation $\prescript{\text{g}}{}{R}_{\text{o}}$ to $\hat{\mathbf{M}}_\text{o}$
    \If{\textbf{S} is True} 
        \State Apply translation $\prescript{\text{g}}{}{t}_{\text{o}}$ to $\hat{\mathbf{M}}_\text{o}$
    \Else
        \State Update $\prescript{\text{g}}{}{t}_{\text{o}}$ and apply to $\hat{\mathbf{M}}_\text{o}$
    \EndIf
    \State Compute intersection $\mathbf{M}_{\text{in}} \gets \overline{\mathbf{M}}_\text{g} \cap \hat{\mathbf{M}}_\text{o}$
    \If{$\mathbf{M}_{\text{in}} \neq \emptyset$}
        \State Find $\mathbf{a}_\text{c}$ and mount to the $\text{c}^{\text{th}}$ candidate
        \State \textbf{S} $\gets$ True
    \Else
        \State \textbf{S} $\gets$ False
    \EndIf
    \State The \acrshort{qp} Solver computes the simulated force.
\EndFor
\end{algorithmic}
\end{algorithm}
\vspace{-1em}
\end{figure}

\subsubsection{Iterative contact localization algorithm}

The deformed mesh of the gripper is reconstructed using the real-time vertex matrix $\overline{\mathbf{V}} \in \mathbb{R}^{n_\text{v} \times 3}$ from the simulation. FoundationPose predicts the transformation matrix from the object frame to the camera frame, denoted as $\prescript{\text{c}}{}{\mathbf{T}}_{\text{o}}$. As the transformation matrix from the camera frame to the gripper jaw frame, $\prescript{\text{g}}{}{\mathbf{T}}_{\text{c}}$, is known, the complete transformation from the object frame to the gripper jaw frame, $\prescript{\text{g}}{}{\mathbf{T}}_{\text{o}}$, can be obtained by chaining $\prescript{\text{g}}{}{\mathbf{T}}_{\text{c}}$ and $\prescript{\text{c}}{}{\mathbf{T}}_{\text{o}}$. This enables the transformation of a copy of the localized object mesh $\mathbf{M}_{\text{o}}$, denoted as $\hat{\mathbf{M}}_{\text{o}}$, into the gripper jaw frame and consequent Boolean intersection computation with respect to the deformed gripper mesh $\overline{\mathbf{M}}_\text{g}$.

When the result of the Boolean intersection, denoted as $\mathbf{M}_{\text{in}}$, is empty, it indicates that there is no contact between the gripper and the object. Otherwise, we identify the vertex $\overline{\mathbf{v}}_\text{j} \in \overline{\mathbf{V}}$ that is closest to the center of $\mathbf{M}_{\text{in}}$. Then, the position of the $j^\text{th}$ vertex in the undeformed state, $\mathbf{v}_\text{j}$, can be retrieved through the undeformed vertex matrix of the gripper jaw, $\mathbf{V}$. Typically, $\mathbf{v}_\text{j}$ lies inside the gripper, but contact can only occur on the inner contact surface. Thus, $\mathbf{v}_\text{j}$ is projected onto the inner contact surface to obtain a point $\mathbf{q}_\text{p}$. This point is used to find the closest contact candidate in the set of candidates $\{ \mathbf{a}_i \}$, denoted as $\mathbf{a}_\text{c}$. The force point actuator is then mounted onto the identified $\text{c}^{\text{th}}$ contact candidate.

Since Foundation pose has a limited frame rate, which significantly affects the system’s performance. To address this issue, we pre-estimate the contact position based on the relative pose between the gripper and the object. In cases where the gripper and the object are not yet in contact, their digital twins are brought into slight contact by adjusting $\prescript{\text{g}}{}{t}_{\text{o}}$, the x-axis translation of $\prescript{\text{g}}{}{\mathbf{T}}_{\text{o}}$, thereby enabling early estimation of the contact position. Since the solver computes the simulated force by minimizing the distance between the position effectors and their targets, the pre-estimation process does not influence the simulation results when no physical contact occurs. $\prescript{\text{g}}{}{t}_{\text{o}}$ is updated accordingly as
\begin{equation}
\label{Eq:cnt_update}
\prescript{\text{g}}{}{t}_{\text{o},x} = l_0 + 0.5l_{\text{bd}} \text{,}
\end{equation}
where $\prescript{\text{g}}{}{t}_{\text{o},x}$ represents the translation along the x-axis. The term $l_0$ denotes the x-distance from the inner contact surface to the origin of the gripper jaw frame in the undeformed state, offset by 1 mm into the surface, and $l_{\text{bd}}$ denotes the width of the object model along the x-axis after applying the predicted rotation, $\prescript{\text{g}}{}{R}_{\text{o}}$. The pseudocode of the proposed iterative contact localization algorithm is provided in Algorithm~\ref{Alg:cntest}.
 Thus, while FoundationPose limits the system to 10 Hz prior to contact, once contact is established and stable, the system runs at 30 Hz, i.e., the frame rate of the RGB-D camera. Overall, the limited frame rate of the contact estimator only constrains the gripper’s approach speed prior to grasp. The frame rate of the \acrshort{ifea} simulation remains sufficient for force control during dynamic grasping of soft objects.

\subsection{Baseline Model Training}
We trained a neural network as a comparative baseline following Zhu et al.\cite{vision_science}. This involved using a ResNet-50 backbone as an image encoder, and feeding a sequence of 10 images into a transformer to predict the grasp force. The dataset consists of grasps of the cylinder and cube used in on-robot evaluation, with three force levels, five motion speeds, and 14 distinct contact locations. The training dataset contained approximately 3.44 hours of video recorded at 30 Hz, which is comparable to that used by Zhu et al.\cite{vision_science}.

\section{Experimental Evaluations, Results, and Discussion}

\subsection{Static Force Evaluation} \label{section: static eval}

\begin{figure}[!t]
\vspace{0.25em}
\centering
\includegraphics[width=0.9\linewidth]{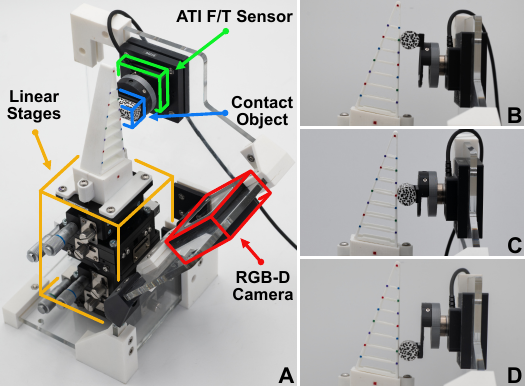}
\caption{Benchtop setup for static evaluation and configurations of contact positions. (A) Experimental setup. (B–D) The gripper is set to contact a 15 mm diameter cylinder at (B) upper, (C) middle, and (D) lower positions}
\label{Fig:bchtop}
\end{figure}

To evaluate the accuracy of the proposed force estimation system, we constructed a benchtop setup and conducted a static evaluation on a single gripper jaw (Fig.\,\ref{Fig:bchtop}A). In this setup, the jaw is mounted on two 3-axis linear stages. The RGB-D camera is mounted at the same initial relative position and viewing angle with respect to the gripper as in the dual-jaw setup (Fig.\,\ref{Fig:kps_demo}A). By preserving the spatial relationship between the camera and the gripper, this benchtop configuration enables us to evaluate the system’s on-robot performance under controlled conditions. As shown in Fig.\,\ref{Fig:bchtop}A, an ATI Nano17 F/T sensor (ATI Industrial Automation, Apex, NC, USA) is placed beneath the grasped object to obtain the ground-truth grasp force ($\mathbf{F}_{\text{gt}}$). The grasped objects were connected to the force sensor via an adapter. The adapter was equipped with a roller bearing that allowed the object to translate along the contact surface. This reduced surface-parallel forces arising from sliding friction during gripper deformations.

The static evaluation was conducted by applying five different load levels to the gripper while interacting with three cylinders of varying diameters: 15, 25, and 35 mm. Each cylinder was tested at three distinct contact positions: upper, middle, and lower (Fig.\,\ref{Fig:bchtop}B-D). The load levels were achieved by displacing the gripper base by 2, 4, 6, 8, and 10 \,mm, with 0 \,mm representing the initial state in which the gripper is positioned but not yet in contact with the object. Because the effective stiffness of the gripper varies across different contact positions, we chose to control the displacement of the linear platform rather than the force applied to the gripper. For each displacement measurement, the gripper is translated to its target displacement. The system is then allowed to settle for 10 seconds to reduce any transient impact and viscous effects. Then the ground-truth and simulated grasp forces ($\mathbf{F}_{\text{sim}}$) are recorded over a subsequent 10 seconds. One cycle is defined as loading the gripper base from its initial position to each displacement level, up to 10 \,mm, and then unloading to each displacement level, until reaching the initial configuration. For each contact condition, five cycles were conducted to ensure the repeatability of the results.

The accuracy of the system was evaluated by computing the \acrfull{rmse} and \acrfull{nrmsd} between the ground-truth and simulated forces. The \acrshort{nrmsd} is calculated by normalizing the \acrshort{rmse} by the observed force range across all loading cycles. The normalizing force range was 11.04\,N in both static force and on-robot evaluation. To quantify the system's error in the worst case, the maximum error of the simulated results during the evaluation was calculated.

The mean \acrshort{rmse} and \acrshort{nrmsd}, along with their standard deviations across five test cycles are shown in Table~\ref{tab: Exp01_result}. The corresponding maximum errors are also reported. We compared errors for both the load and unload phases under different contact and displacement conditions (Fig.\,\ref{Fig:exp1_result}A–C). It can be observed that when contact occurs at the upper or lower positions, the estimated grasp force becomes less accurate as the cylinder diameter increases, evidenced by the data points deviating further from the gray dashed unity line. In contrast, the estimation accuracy remains consistent as the cylinder diameter changes at the middle contact position, with a mean \acrshort{rmse} ranging from 0.24 to 0.28\,N and a mean \acrshort{nrmsd} ranging from 2.16 to 2.54\,\% (Table \ref{tab: Exp01_result}). Our results reveal two factors that affect accuracy: (1) the amount of internal strain, and (2) the occlusion of key points.

\begin{table}[!t]
    \centering
    \caption{Simulated grasp force accuracy under different contact positions and cylinder diameters.}
    \renewcommand{\arraystretch}{1}
    {
    \begin{tabular}{ccccc}
        \toprule
        \makecell{\textbf{Contact} \\ \textbf{Position}} & \makecell{\textbf{Ø}(mm)} & \makecell{\textbf{\acrshort{rmse}}(N)} & \makecell{\textbf{\acrshort{nrmsd}}(\%)} & \makecell{\textbf{Max} \textbf{Error}(N)}\\
        \midrule
        \multirow{3}{*}{Upper} & 15 & 0.22 $\pm$ 0.03 & 1.97 $\pm$ 0.30 & 0.54 \\
                            \cmidrule{2-5}
                            & 25 & 0.52 $\pm$ 0.09 & 4.67 $\pm$ 0.79 & 1.02 \\
                            \cmidrule{2-5}
                            & 35 & 0.57 $\pm$ 0.05 & 5.20 $\pm$ 0.50 & 1.52 \\
        \midrule
        \multirow{3}{*}{Middle} & 15 & 0.24 $\pm$ 0.02 & 2.16 $\pm$ 0.21 & 0.62 \\
                             \cmidrule{2-5}
                             & 25 & 0.28 $\pm$ 0.07 & 2.54 $\pm$ 0.59 & 0.98\ \\
                            \cmidrule{2-5}
                            & 35 & 0.24 $\pm$ 0.07 & 2.18 $\pm$ 0.60 & 0.60 \\
        \midrule
        \multirow{3}{*}{Lower} & 15 & 0.69 $\pm$ 0.05 & 6.26 $\pm$ 0.48 & 1.55 \\
                            \cmidrule{2-5}
                            & 25 & 0.62 $\pm$ 0.11 & 5.62 $\pm$ 0.98 & 1.47 \\
                            \cmidrule{2-5}
                            & 35 & 1.21 $\pm$ 0.15 & 10.93 $\pm$ 1.37 & 3.33 \\
        \bottomrule
    \end{tabular}
    }
    \label{tab: Exp01_result}
\end{table}

\subsubsection{Effect of Internal Strain}\label{internal_strain_effect}
Given its linear elasticity assumption, our system performance depends on the degree of nonlinear material behavior. This is primarily induced by the increased gripper internal strain. During the deformation, the gripper experienced different ratios of internal strain and shape transformation due to variations in effective stiffness across contact positions. At the lower contact position (Fig.\,\ref{Fig:exp1_result}A), where the gripper exhibited higher effective stiffness compared to the other two positions, our system tended to underestimate the grasp force as the object size increased under large displacements. This is because in the regions with higher effective stiffness, the gripper has less observable shape transformation but more unobservable internal strain.

\subsubsection{Effect of Key Point Occlusion}
\begin{figure}[t!]
    \vspace{0.25em}
    \centering
    \includegraphics[width=\linewidth]{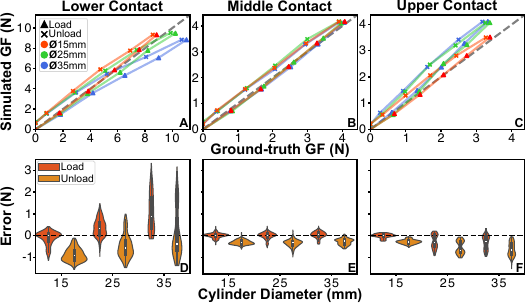}
    \caption{Contributions of contact position and cylinder size to grasp force estimation error. Top row (A–C): Comparison between $\mathbf{F}_{\text{sim}}$ and $\mathbf{F}_{\text{gt}}$ during load and unload phases for cylinders with diameters of 15, 25, and 35 mm, under three contact positions: (A) lower, (B) middle, and (C) upper. Bottom row (D–F): Violin plots of estimation error across cylinder diameters for contact positions: (D) lower, (E) middle, and (F) upper. (D) and (F) indicate a bimodal error pattern, in which our system tends to underestimate the force at lower contact position and overestimate it at upper position. (A) and (C) further demonstrate that these estimation biases become more pronounced under larger deformations.}
    \label{Fig:exp1_result}
\end{figure}

The accuracy of our system also depends on the visibility of the key points. This is primarily affected by the size of the object being grasped, as larger objects tend to occlude more key points near the contact region. This effect was amplified at the upper contact position. At this position, our system tended to overestimate the grasp force as the cylinder diameter increased (Fig.\,\ref{Fig:exp1_result}C). This is because the outer position effectors were occluded, thus reducing the contribution of the upper-region constraints in the simulation. Therefore, the solver accounts for more of the lower-region constraints, where the gripper exhibits a higher effective stiffness, leading to an overestimation of the grasp force. The results at the middle contact position, where only inner effectors tended to be occluded as object diameter increased, suggest that the influence of inner effector occlusions is relatively limited, as the estimation accuracy remains consistent with increasing cylinder size. The effect of occlusions is less pronounced at the lower contact position, where the effect of internal strain becomes dominant.

Due to the position-dependent coupling between the effects of internal strain and key point occlusion that affect accuracy, we note that defining a quantitative metric relating accuracy purely to visibility or object geometry is difficult. We further analyzed the error distribution in different load phases, as shown in Fig.\,\ref{Fig:exp1_result}D–F. The results indicate that the estimation error was generally lower in the load phase than in the unload phase. This observed hysteresis suggests an unmodeled viscous effect. When contact occurred at the upper or lower positions, the error distribution exhibited a bimodal pattern. The diminished accuracy at these two contact positions both occur under high deformation, but are attributable to two distinct factors, which are the occlusion of outer key points in the upper contact scenario, and more unobservable internal strain in the lower contact scenario.

\subsection{Contact Estimator Evaluation}

In the static force evaluation, the contact position in the simulation was predefined, whereas in real-world applications, it is typically unknown. Using our iterative contact localization algorithm, the system can adapt to different objects and remain effective even when the contact position shifts during grasping. In this experiment, we aimed to evaluate the necessity of the contact estimator for the overall system performance. Due to the difficulty in directly quantifying the accuracy of the estimated occluded contact position, we assess its accuracy by analyzing the improvement in grasp force estimation performance. To control for the error in mesh quality and object pose estimation attributable to FoundationPose, we used a CAD model as the digital twin of the cylinder and hard-coded the its pose by initializing the digital twin to the known pre-contact state. The translation along the x-axis was then acquired by computing the displacement of the reference point from its initial position. 

\begin{table}[!t]
    \centering
    \caption{Grasp force estimation accuracy (\acrshort{rmse} (N) $\pm$ \acrshort{std}) under different actuator positions and the contact estimator.}
    \renewcommand{\arraystretch}{1}
    \resizebox{0.48\textwidth}{!}{
    \begin{tabular}{ccccc}
        \toprule
          \diagbox[width=5em]{\textbf{Real}}{\textbf{Sim}}  & \makecell{Contact Estimator} & Upper & Middle & Lower \\
        \midrule
        Upper & \cellcolor{ce}0.20 $\pm$ 0.04 & \cellcolor{gt}0.22 $\pm$ 0.03 & 1.21 $\pm$ 0.10 & 5.91 $\pm$ 0.13 \\
        \midrule
        Middle &  \cellcolor{ce}0.26 $\pm$ 0.02 & 0.93 $\pm$ 0.14 & \cellcolor{gt} 0.24 $\pm$ 0.02  & 4.93 $\pm$ 0.17 \\
        \midrule
        Lower & \cellcolor{ce} 0.61 $\pm$ 0.12 & 4.40 $\pm$ 0.52 &  3.61 $\pm$ 0.53 & \cellcolor{gt} 0.69 $\pm$ 0.05 \\
        \bottomrule
        \addlinespace[0.5mm]        
        \multicolumn{5}{p{\linewidth}}{\footnotesize\textsuperscript{*} Blue cells indicate cases where the preset contact position matched the actual contact position, while green cells indicate cases where the system was implemented with the contact estimator.}\\
    \end{tabular}
    }
    \label{tab: Exp02_result}
\end{table}

We repeated the same experimental procedure described in the static evaluation, with the gripper interacting with a 15 mm cylinder at three distinct contact positions. For each contact position, the simulation actuator was configured in four different ways: (1) fixed on the upper position, (2) fixed on the middle position, (3) fixed on the lower position, and (4) initialized at the middle position with the contact estimator activated, i.e., our method. The mean \acrshort{rmse} and its standard deviation are reported in Table~\ref{tab: Exp02_result}. The results indicate that the contact estimator is essential to the system. Incorrect fixed actuator configurations significantly degraded estimation accuracy, whereas the contact estimator achieved similar accuracy to fixed configurations where the actuator position is correctly predefined.

\subsection{On-robot Grasp Force Evaluation}

\begin{figure*}[!t]
\vspace{0.25em}
    \centering
    \includegraphics[width=0.97\linewidth]{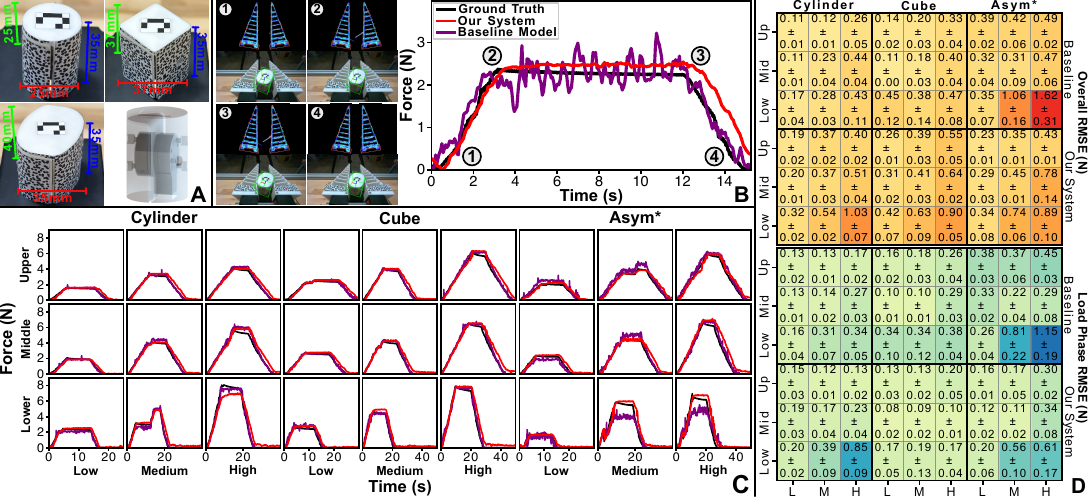}
     \caption{Objects with built-in load cells and experimental results of the on-robot evaluation. (A) Cylinder, cube, asymmetric object with built-in load cells and CAD design under perspective view. (B) Estimation results when grasping the asymmetric object at lower position under low force level. Images of the physical grippers and their digital twins at four stages of the grasping process are shown. (C) Representative grasp force estimation results across different grasping conditions. \textbf{Asym$^{*}$} indicates that the asymmetric object is unseen for both our system and baseline model. (D) Mean RMSEs along with STDs of the entire grasping process and the load phase for both our system and baseline model. L, M, and H represent low, medium, and high force level.}
    \label{Fig: Exp3data}
\end{figure*}

To evaluate the system’s performance in an on-robot scenario, we conducted grasping tests on three commonly encountered object shapes. The dual-jaw gripper was mounted on a Kinova Gen3 7-\acrshort{dof} robotic arm (Kinova, Quebec, Canada). The test objects, a cylinder, a cube, and an asymmetric object, were 3D-printed as two halves, so that they could contain an embedded F/T sensor for $\mathbf{F}_{\text{gt}}$ measurement (Fig.\,\ref{Fig: Exp3data}A). Using our mesh reconstruction approach, the reconstructed meshes achieve L2 Chamfer distances of 2.36, 3.53, and 3.35\,mm with respect to their corresponding CAD models for the tested objects. In comparison, the raw meshes exhibit substantially larger Chamfer distances of 13.39, 14.69, and 15.69\,mm, while the coarsely calibrated meshes yield distances of 2.80, 6.69, and 3.26\,mm. The large distance observed for the coarsely calibrated cube mesh highlights the necessity of our method.

The robot was controlled to grasp each object at the upper, middle, and lower contact positions with a grasping speed of 1 mm/s. To further investigate the system’s performance under varying force levels (low, medium, and high), the gripper was actuated with incremental control currents of 300 mA, 500 mA, and 800 mA. To equilibrate the transient effects, the gripper stayed fully grasped for six seconds in each grasping process. Each force level was repeated five times. We recorded the videos of each grasping process and benchmarked our system by running the baseline model offline. We also evaluated the system’s performance in scenarios where the contact position shifts during the grasping process by grasping the asymmetric object.

Embedding the F/T sensor within the object allowed us to measure the ground-truth force without interfering with the natural grasping process. To ensure reliable ground-truth measurements while minimizing the impact on our system bandwidth, we used a marker-based optical tracker, Micron Tracker 4 (Claronav, Toronto, Canada), to establish the kinematic chain from the sensor frame to the gripper jaw frame, enabling the transformation of the sensor-measured force for validation purposes. Two optical tracking markers were attached to both the top surface of the object and the RGB-D camera mount. The estimated grasp and manipulation force, $\mathbf{F}^{g}_{\text{sim}}$ and $\mathbf{F}^{m}_{\text{sim}}$, were calculated using the method described by Yoshikawa and Nagai \cite{GFdefinition},
\begin{equation}\label{Eq: Definition of GF}
    \begin{bmatrix}
        \ \mathbf{F}^{g}_{\text{sim}} \ \\
        \ \mathbf{F}^{m}_{\text{sim}} \
    \end{bmatrix} = \begin{bmatrix}
        \ \min(\lvert{\mathbf{F}_{\text{gl}}} \rvert, \lvert {\mathbf{F}_{\text{gr}}} \rvert) \ \\
        \ \mathbf{F}_{\text{gl}}-\mathbf{F}_{\text{gr}} \
    \end{bmatrix}
\end{equation}
where $\mathbf{F}_{\text{gl}}$ and $\mathbf{F}_{\text{gr}}$ denote the estimated force on \acrshort{gl} and \acrshort{gr}, respectively and the postive direction of $\mathbf{F}^{m}_{\text{sim}}$ aligns with that of $\mathbf{F}_{\text{gl}}$. Due to the difficulty of obtaining reliable measurements of the ground-truth manipulation force, we only evaluated the grasp force in this experiment.

We categorized the grasping process into four stages (Fig.\,\ref{Fig: Exp3data}B). At Stage 1, the grippers begin grasping the object. At Stage 2, the full commanded grasp is achieved. At Stage 3, the gripper begins to release the object. At Stage 4, the object is fully released. The load phase is defined as the period between Stage 1 and 2. The mean \acrshort{rmse}s and standard deviations for the entire grasping process and the load phase were calculated separately for both the baseline model and our system (Fig.\,\ref{Fig: Exp3data}D). Except for the worst-case scenario, where the gripper was commanded to grasp the object with a high force level at the lower contact position, the overall \acrshort{rmse} of our system ranged from 0.19 to 0.78 \,N, and the load phase \acrshort{rmse} ranged from 0.08 to 0.56 \,N with high repeatability. The corresponding \acrshort{nrmsd} ranges from 1.75 to 7.04 \% over the entire grasp process and from 0.71 to 5.08 \% during the load phase. Due to the 30\,Hz effective frame rate limitation, there was a 0.25\,s delay before the onset of force. The results reveal a clear pattern, in which the \acrshort{rmse} increased with higher force levels due to the combined effects of observing fewer outer key points, and the increasing internal strain. In addition, the results demonstrate the robustness of the proposed contact localization algorithm when using a reconstructed mesh.

Both the baseline model and our system achieved high accuracy when grasping the cylinder and cube, which are included in their training datasets (this corresponds to the \acrshort{dlc} dataset for our system). However, when grasping the asymmetric object, the baseline model exhibits a substantial drop in accuracy, while our system maintains high performance (Fig.~\ref{Fig: Exp3data}B-D). This result demonstrates the generalization capability of the data-driven method and the robustness of our key point acquisition pipeline. It also indicates that our system can accurately estimate the grasp force even when the contact positions shift during the grasping process. Our system's errors primarily occurred between Stages 2 and 4 of the grasping process and were likely due to the nonlinear behavior and viscosity of the soft material.

\subsection{On-robot Manipulation Force Evaluation}

To evaluate manipulation force estimation accuracy, the gripper was commanded to stably grasp a 100\,g calibration weight (0.98\,N). It then rotated counter-clockwise 360$^\circ$, and then reversed back to its initial pose. Due to gravity, the manipulation force will vary with the gripper's rotational angle. As illumination also changes during rotation, we conducted comparative experiments with an external light source affixed to the manipulator to analyze the effect of lighting conditions on system accuracy. Both conditions were repeated five times to ensure repeatability. The \acrshort{rmse} is 0.12 $\pm$ 0.00\,N with an external light source, compared to 0.28 $\pm$ 0.02\,N without one. The close tracking of the polar manipulation force (Fig.~\ref{Fig:mf_and_chips}A) demonstrates the system's accuracy in estimating the manipulation force under stable lighting conditions. The results also indicate that our system maintained accurate force estimation except under severe low-light conditions (90$^\circ$–270$^\circ$), where the flipped gripper occluded environmental light. The increased error is mainly attributed to the degraded point cloud quality, with reduced inference confidence of \acrshort{dlc} also contributing secondarily.

\subsection{Grasping a Potato Chip}

\begin{figure}[!t]
\vspace{0.25em}
\centering
\includegraphics[width=\linewidth]{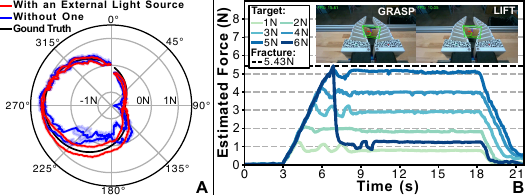}
\caption{Manipulation force evaluation and potato chip grasping results. (A) Force trajectory in polar coordinates, where the angle represents the gripper rotation. The shaded region indicates the standard deviation. (B) System performance with force control during potato chip grasping and lifting.
}
\label{Fig:mf_and_chips}
\end{figure}

To demonstrate the capability of our system for force control during delicate grasping, we commanded the gripper to repeatedly grasp and lift a potato chip at a closing speed of 4\,mm/s while incrementally increasing the force threshold in steps of 1\,N until fracture occurred. The gripper successfully grasped and lifted the potato chip at a force level of 5\,N, whereas the chip fractured at 5.43~N (Fig.\,\ref{Fig:mf_and_chips}B), demonstrating the system's capability to perform delicate grasping.

\section{Conclusion and Future Work}

In this letter, we presented the design and evaluation of a model-based visual grasp force sensing approach for deformable fin-ray-shaped grippers. This approach comprised real-time \acrshort{ifea} simulations and a learning-based key point acquisition pipeline using a RGB-D camera. A novel mesh reconstruction pipeline and model-based iterative contact localization algorithm were developed, enabling reliable contact estimation when grasping unseen objects of varying shapes under occlusion. In these tests, our approach outperformed an end-to-end deep learning baseline model and maintained performance under moderately bright lighting conditions. We further demonstrated our system's dynamic performance by using force control to grasp a potato chip.

Our approach is likely compatible with other passive compliant grippers that exhibit distinct shape transformations under external loads. In the future, we plan to investigate gripper-specific adjustments to the acquisition pipeline and the \acrshort{ifea} simulation parameters to validate such compatibility. Currently, the simulation models materials as linearly elastic, which limits the gripper material to those with adequately linear responses over the target sensing range. This facilitates observability of internal strain, which is crucial for maintaining high accuracy. Moreover, material aging effects are not considered in our framework. We will address these in the future by investigating alternative time-aware modeling approaches.

\addtolength{\textheight}{-12cm}
                                  
\bibliographystyle{IEEEtran}
\bibliography{references}

\end{document}